American Sign Language
Linguistic Research Project

http://www.bu.edu/asllrp/

# Challenges for Linguistically-Driven Computer-Based Sign Recognition from Continuous Signing for American Sign Language *

**Carol Neidle**

Boston University, Boston, MA

## Abstract

There have been recent advances in computer-based recognition of isolated, citation-form signs from video. There are many challenges for such a task, not least the naturally occurring inter- and intra- signer synchronic variation in sign production, including sociolinguistic variation in the realization of certain signs. However, there are several significant factors that make recognition of signs from continuous signing an even more difficult problem. This article presents an overview of such challenges, based in part on findings from a large corpus of linguistically annotated video data for American Sign Language (ASL). Some linguistic regularities in the structure of signs that can boost handshape and sign recognition are also discussed.

## Introduction

There have been recent advances in computer-based recognition of isolated, citation-form signs from video (e.g., De Coster*, et al.* 2020; Rastgoo*, et al.* 2020; Jiang*, et al.* 2021; Dafnis*, et al.* 2022a; Dafnis*, et al.* 2022b). There are many challenges for such a task, not least the naturally occurring inter- and intra- signer synchronic variation in sign production, including sociolinguistic variation in the realization of certain signs; for an overview and relevant references, see, e.g., Section 8.2 in Brentari (1995). However, there are several significant factors that make recognition of signs from continuous signing an even more difficult problem.

## 1. Overview of Differences between Continuous Signing & Isolated Sign Production

First, in natural signing, the duration and articulation of signs tend to be reduced, to varying degrees (Tyrone & Mauk 2010). There are also some phonological modifications argued by Napoli, *et al.* (2014) and Napoli & Liapis (2019) to be motivated by articulatory effort reduction. Examples from Napoli & Liapis (p. 35) include the following, as well as some of the phenomena discussed in Section 4:[1]

• **Weak freeze**: wherein a sign normally produced through symmetric movement of both hands is produced with the non-dominant hand (i.e., the "weak hand") held fixed, without movement (first described by Padden & Perlmutter (1987)). For example, in **Figure 1**, the sign INTERPRET is shown on the left as it is normally produced, with symmetrical alternating movement of both hands. However, in the second production, only the dominant hand moves; the non-dominant hand remains fixed.

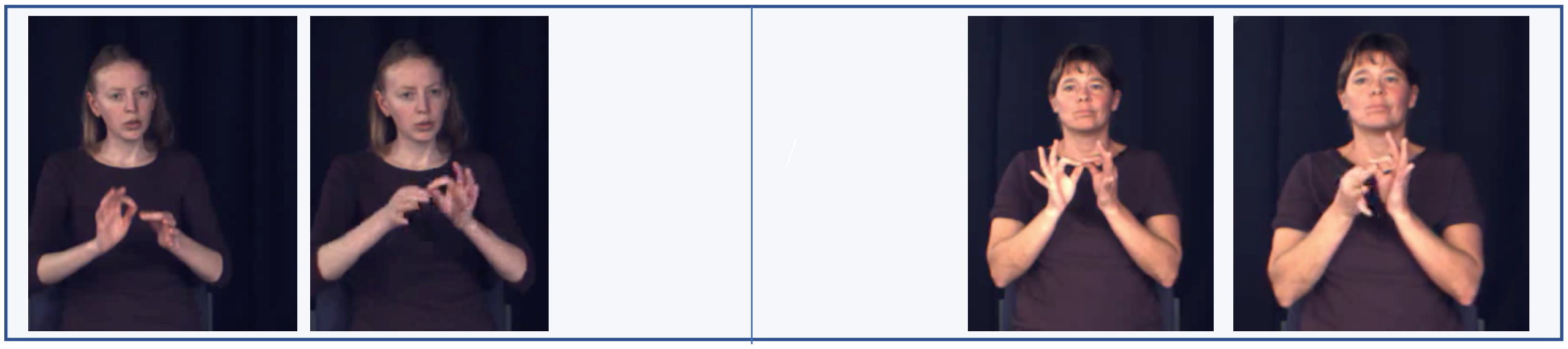

**Figure 1**. The sign INTERPRET, from the ASLLRP Sign Bank (ASLLRP 2017-2023)[2] with "weak freeze" on the right: the non-dominant hand does not move.

• **Iteration loss**: where a sign is produced with fewer repetitive movements than normal.

• **Distalization**:[3] where the joint normally used for the sign is replaced by movement of a joint more distal from the torso, e.g., a sign normally produced with flexion at the elbow is instead produced with flexion at the wrist. An example is shown in **Figure 2,** where the sign for REALLY is normally produced with flexion of the elbow, but is shown on the right with flexion of the wrist instead (and a reduced movement). For further details about possible joint involvement, see Napoli, *et al.* (2011).

---

[1] See also Napoli, *et al.* (2014). Video examples illustrating weak freeze, weak drop (see Section 4.6), distalization, and lowering are linked into Anderson (2022) Section 4.1 at <https://ecampusontario.pressbooks.pub/essentialsoflinguistics2/chapter/4-10-signed-language-phonology/>**.**

[2] The ASLLRP Sign Bank is accessible from <https://dai.cs.rutgers.edu>.

[3] A greater tendency for distalization has been observed in the signing of those with Parkinson's disease (Poizner & Kegl 1992), in whispering, and also in gay signing (Blau 2017). There are also some differences in the frequency of distalization found in those acquiring ASL (Mirus*, et al.* 2000; Meier 2005).

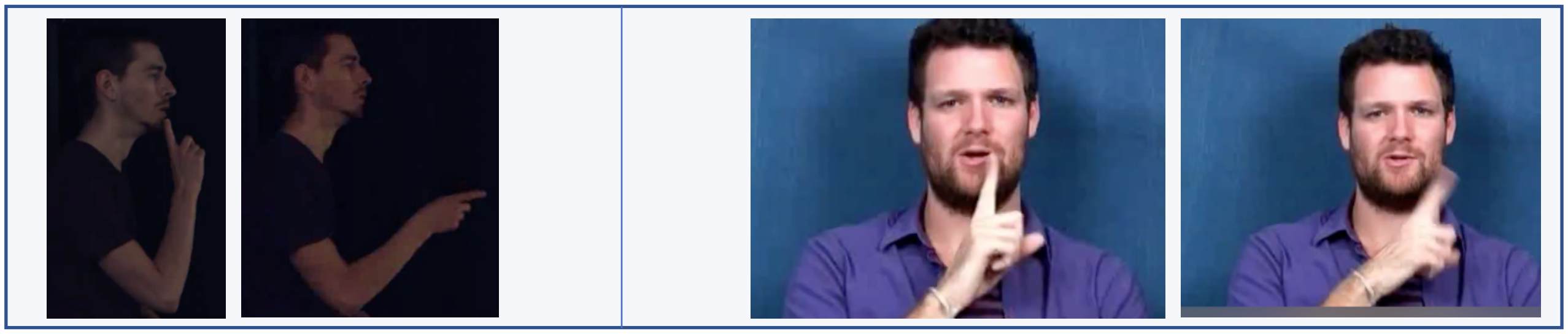

**Figure 2**. Distalization (right) of the sign REALLY, from the ASLLRP Sign Bank (ASLLRP 2017-2023)

• **Joint freeze**: where a sign normally produced with flexion at more than one joint is produced without flexion at one of those joints; e.g., the sign glossed as HURRY or RUSH, normally produced with flexion at both the wrist and the elbow (as shown in the first illustration **Figure 3**), can be produced with flexion only at the wrist (as shown in the second).

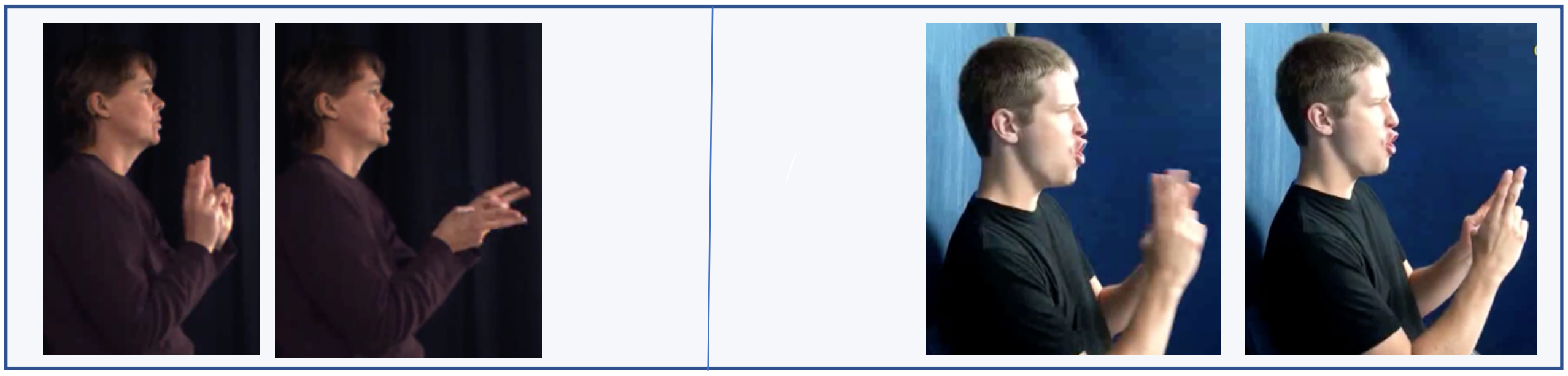

**Figure 3**. HURRY, from the ASLLRP Sign Bank (ASLLRP 2017-2023), with joint freeze on the right

Another phenomenon that may occur, resulting in reduction of effort, is:

• **Lowering**: where the place of articulation of a sign normally produced high on the body may be lowered, as when the sign KNOW is produced with contact on the upper cheek rather than the forehead. Compare the two examples of the sign KNOW shown in **Figure 4**.

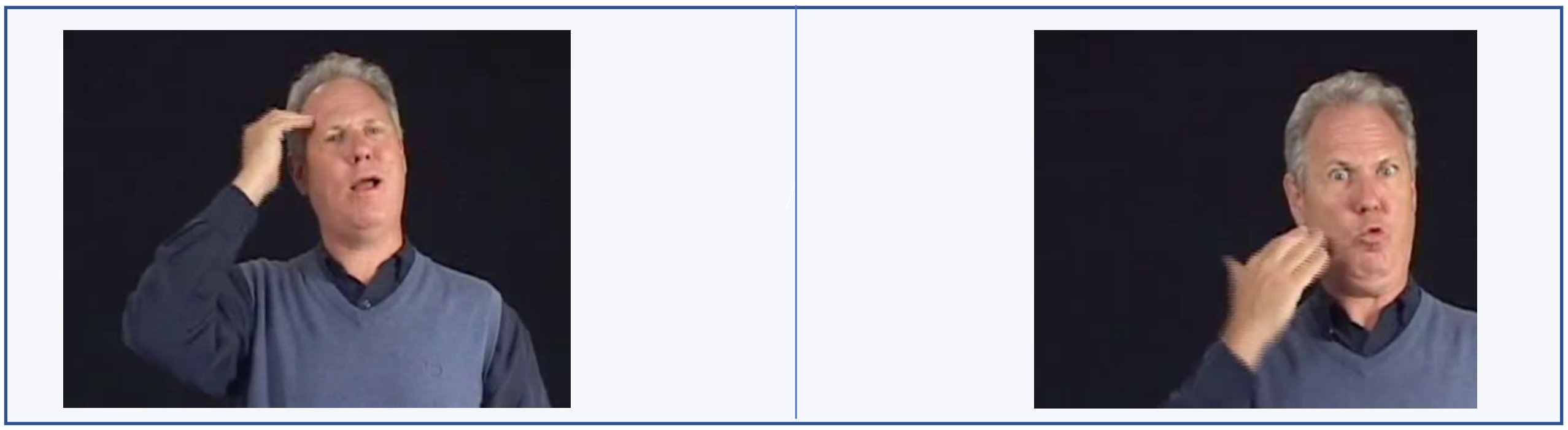

**Figure 4**. KNOW, from the ASLLRP Sign Bank (ASLLRP 2017-2023), with lowering on the right

Second, sign production can be affected by neighboring signs, with respect to handshape, orientation, and/or location; this phenomenon is referred to as coarticulation and is addressed in Sections 2 and 3.

An additional set of complications relates to the way in which the two hands interact in sequences of 1- and 2-handed signs; an overview of these issues is presented in Section 4. However, there are also linguistic constraints on the internal structure of signs that can be leveraged to boost sign recognition, as discussed in Sections 5 and 6.

## 2. Coarticulation

### 2.1. Introduction

Coarticulation, a term dating back to 1933,[4] was initially defined and studied in relation to spoken languages. It refers to cases in which the acoustic and articulatory properties of a speech segment are modified based on those of an adjacent segment. This phenomenon is also found in signed languages, although, of course, manifested in ways specific to the articulatory properties of the visual-gestural modality. Accordingly, coarticulation in signed languages potentially involves modifications in hand shape, orientation, and/or location, on one or both hands. Thus, there are many ways in which the appearance of a given sign can vary based on context, which poses a challenge for computer-based sign recognition from continuous signing videos. However, as compared with spoken language, coarticulation seems overall to be less prevalent, although coarticulation effects in signed languages usually spread over longer durations (Benner 2021).

### 2.2. Location

Several researchers have compared the long-distance coarticulation effects in ASL with respect to location (cases where the location of a sign may be affected by the location of a sign in an adjacent syllable (Cheek 2001a; Lucas*, et al.* 2002; Mauk 2003; Tyrone & Mauk 2010; Russell*, et al.* 2011; Mauk & Tyrone 2012)) to Vowel-to-Vowel coarticulation in spoken languages, whereby the quality of the vowel in one syllable may be adjusted in its articulation to become more similar to a vowel in an adjacent syllable (Whalen 1990; Farnetani 1997; Cole*, et al.* 2010). In both signed and spoken languages, these effects are somewhat dependent on the rate of signing.

However, although these effects vary among individuals for both signed and spoken languages, Grosvald & Corina (2012) and Tyrone & Mauk (2010), among others, have found that these coarticulation effects are overall much weaker in signed than in spoken language.

### 2.3. Handshape

There are few studies in the literature of handshape coarticulation in signed languages. Handshape coarticulation has been most studied specifically for fingerspelling, where a maximum amount of handshape coarticulation occurs (e.g., Wilcox 1992; Jerde*, et al.* 2003; Channer 2010; Keane*, et al.* 2012). Regressive handshape assimilation also occurs quite productively at morpheme boundaries within compounds (see, e.g., Liddell & Johnson (1986), Sandler (1993, 2009)). When handshape coarticulation occurs, the handshape may totally assimilate to the neighboring sign's handshape (as often occurs within compounds, for example), or, as happens quite most often at boundaries between signs, it can assume a position intermediate, in some respect, between the canonical handshape for the sign in question and the adjacent handshape.

[4] See Hardcastle & Hewlett, eds. (1999), *Coarticulation: Theory, data and techniques* for history and further details.

Prior studies of handshape coarticulation have generally focused on a few signs and/or handshapes. For example, Cheek (2001b) focused on the '1' 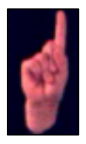 and the '5' 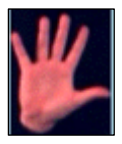 handshapes occurring in adjacent signs. Her explanation for selecting those specific handshapes was the following: "they are two of the six most frequently occurring handshapes in ASL" and "they are not easily confused." She found that, as in spoken language, the rate of signing was a factor in the occurrence of coarticulation.

The observation that handshape assimilation is quite limited, with the '1' handshape particularly prone to assimilation, goes back to Liddell & Johnson (1989), p. 250:

> There are numerous instances of assimilation in ASL. For example, the hand configuration of the sign ME typically assimilates to that of a contiguous predicate in the same clause... The extent to which signs other than ME assimilate to the hand configuration of another sign... appears to be considerably more limited.

**Figure 5** shows an example of handshape assimilation of this same 1st-person index pronoun, which we gloss as IX-1p, from our ASLLRP SignStream® 3 corpus (Neidle, *et al.* 2022a, 2022b), accessible from <https://dai.cs.rutgers.edu/dai/s/dai>. This sign is normally produced with the '1' handshape, as shown in (A) (from Cory_2013-6-27_sc110.ss3, Utterance 17). However, in the sequence of signs shown in (B), (C), and (D) (from Cory_2013-6-27_sc107.ss3, Utterance 10), the handshape used for this same sign in (C) assimilates to the handshape that will be used in the following sign, shown in (D).

Bayley, *et al.* (2002) discuss the wide range of sources of variation in the realization of the '1' handshape—which include, but are not limited to, effects pf coarticulation. They found that the '1' handshape was quite often realized by a different handshape, as a result of various factors. They found, furthermore, that the likelihood of coarticulation with the '1' handshape was, in large part, dependent on its grammatical function, with $1^{st}$ and $2^{nd}$ person pronouns more likely than other similar signs to undergo coarticulation.

Fenlon, *et al.* (2013) studied "variation in handshape and orientation in British Sign Language" (BSL) of the '1' hand configuration, comparing the findings for BSL with Bayley, *et al.*'s findings for ASL. Fenlon, *et al.* likewise found a very high frequency of signs with the '1' handshape realized with a different handshape. The factors that influenced the likelihood of a different handshape occurring were, in order of importance, the preceding handshape, the following handshape, grammatical category and indexicality, and lexical frequency.

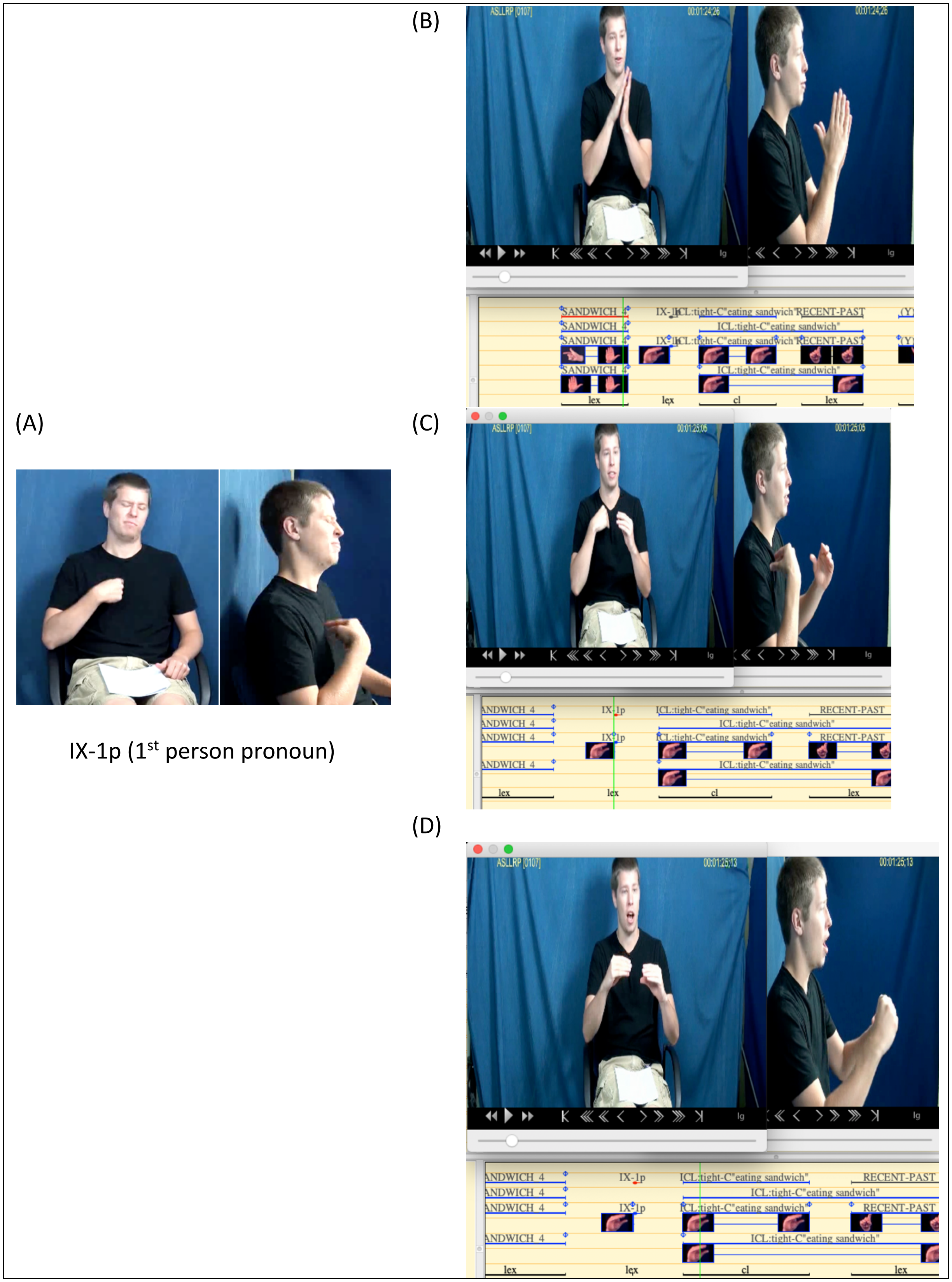

**Figure 5**. Handshape coarticulation: IX-1p articulated in (C) with the handshape of sign in (D)

McKee (2011) looked specifically at New Zealand Sign Language (NZSL) sign produced with both palms facing upward. This is very similar to the ASL indefinite particle sign that we gloss as “part:indef”. The handshape from the previous sign frequently results in modification of the typical palms-up ‘5’ handshape. Interestingly, this also occurs frequently in ASL with this sign, as shown in **Figure 6**.

Consider the final two signs in this sentence (Cory_2013-6-27_sc107, Utterance 52) from the ASLLRP SignStream® 3 corpus: WHEN, articulated with the ‘1’ handshape on both hands, and the indefinite particle (part:indef):

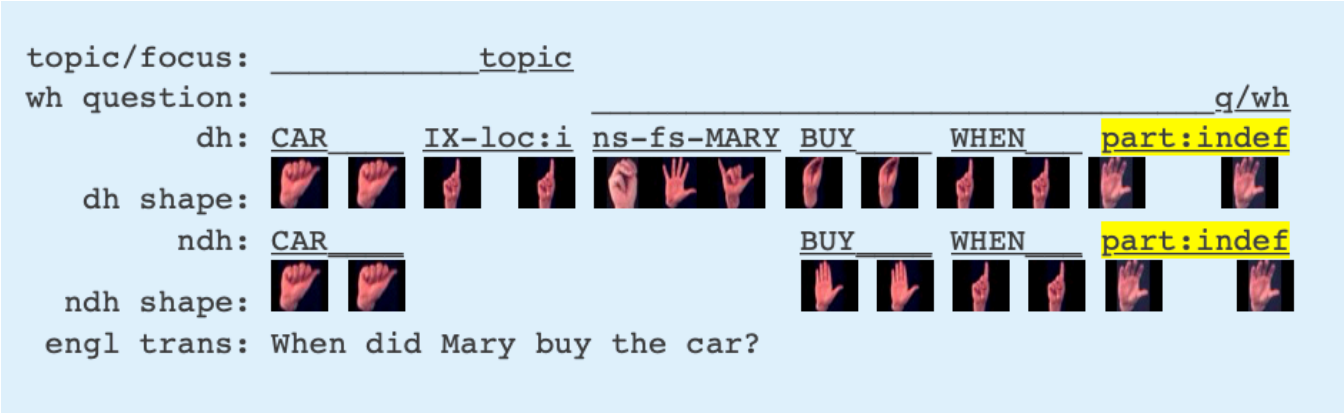

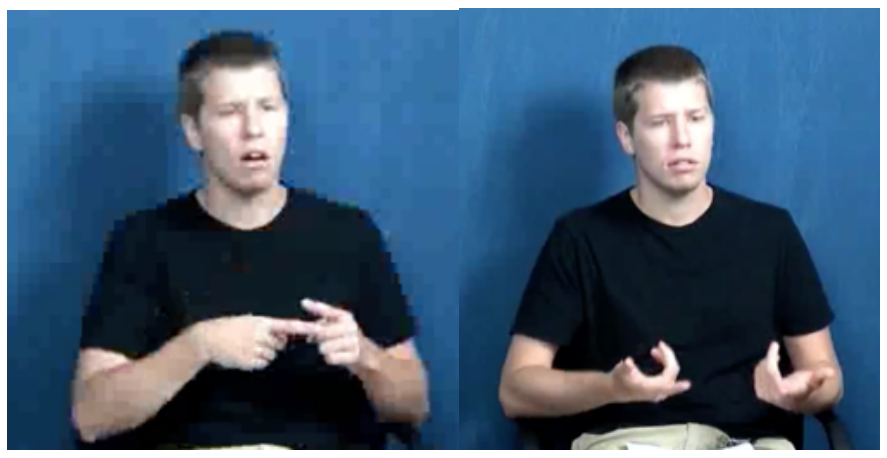

WHEN part:indef

The typical handshape for part:indef, shown below, is modified in the above example, where the index finger from the previous sign remains selected.

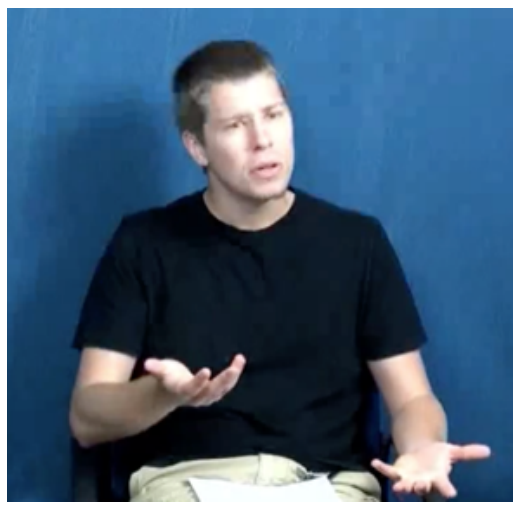

(from Cory_2013-6-27_sc107, Utterance 91)

**Figure 6.** Handshape coarticulation with the ASL sign "part:indef"

Ormel, *et al.* (2017) looked at “Coarticulation of Handshape in Sign Language of the Netherlands” (NGT), with a focus on the articulation of the thumb in flat handshapes (B handshapes). They looked specifically at the thumb position as an indicator of coarticulation. They found (pp. 14-15):

> The realization of the preceding and following signs influenced the realization of the B handshape in target signs in NGT. This turned out to be true for the degree of flexion of the thumb depending on the degree of flexion of the preceding or following signs, as well as for the degree of abduction of the thumb, which similarly depended on the degree of abduction of the preceding or following signs.

Thus, previous studies have focused on handshape coarticulation in various signed languages in relation to a limited number of handshapes, and they have also found that it occurs more frequently with certain specific types of signs. Previous research has also shown that the likelihood of coarticulation generally increases with rate of signing.

## 3. Patterns & Prevalence of Handshape Coarticulation in our Datasets

Since handshape coarticulation could potentially complicate computer-based sign recognition of segmented signs from continuous signing that relies on start and end handshape identification, we wanted to determine the extent to which this normally occurs in continuous signing in ASL.

To get a sense of the prevalence of handshape coarticulation in continuous signing, we explored several datasets distributed through the American Sign Language Linguistic Research Project (ASLLRP)—segmented from continuous signing videos, based on the manual annotations carried out using SignStream® (Neidle 2002a, 2002b, 2007; ASLLRP 2017-2022; Neidle*, et al.* 2018; Neidle 2022) available as of December 2022, in which the start and end frames of each sign had been identified. This included 2,124 ASL sentences collected from native signers at Boston University as well as 527 example sentences intended for the all-ASL dictionary currently in the final stages of development by DawnSignPress (DSP) and contributed by DSP. We excluded from consideration classifiers, gestures, and fingerspelled signs. This expanding collection of linguistically annotated video data is available for viewing, browsing, searching, and downloading from <https://dai.cs.rutgers.edu/dai/s/dai>.

There are several lexical items previously identified in the literature (as discussed in Section 2.3) as extremely prone to handshape coarticulation, such as index signs (pronouns, determiners, adverbials), which also undergo coarticulation in our data set with a very high frequency, with adjustments to handshape based on the preceding and/or following sign. These signs are also brief in duration, often corresponding to 1 or 2 frames in 30 frame-per-second videos. In addition, the sign that we have glossed as "part:indef" (Conlin*, et al.* 2003), which, we have argued, functions as an indefinite focus marker, articulated with palms up and moving outward (normally 2-handed, but sometimes produced with a single hand) is also extremely prone to undergoing coarticulation, with handshape modified by the handshape of the preceding sign. As discussed on p. 7, a similar sign in New Zealand Sign Language was likewise identified as frequently undergoing coarticulation (McKee 2011).

In our recent research, we set out to investigate the prevalence of handshape coarticulation specifically (not including other types of coarticulation, e.g., with respect to location, orientation, etc.). We examined a random sample of the approximately 75% of the total number of segmented sign clips we had—excluding pronouns and other index signs as well as part:indef, in the two ASLLRP collections of continuous signing videos. There is a fair amount of variation in handshape configuration that is found normally in citation-form signs. We set out to identify cases where the handshape variation that occurred at the start and/or end of a given segmented sign clip was not explicable in terms of such variation, but rather, where it was affected, to differing degrees, by the handshape of the preceding or following sign. The total number of signs where the start and/or end handshape on the dominant and/or non-dominant hand was modified, to some visibly noticeable degree, by the preceding or following sign's handshape, was only 158, out of the total number of signs examined, 11,077 – which comes to a frequency of less than 1.5%. Thus, the first obvious conclusion is that handshape coarticulation is extremely limited, outside of the few lexical items with which it occurs particularly productively.

Furthermore, of those 158 examples, about 56 (i.e., just over a third of them) had effects that were quite subtle, with the handshape modified only slightly, by subtle changes in the spread of the fingers, curvature of the hand, or selected fingers, influenced by the preceding or following handshape; thus, these are cases where the changes would be unlikely to interfere with computer-based recognition of the sign based on hand configuration. In the example shown in **Figure 7**, the end handshape of the sign FUTURE starts to transition to the following handshape before the end of the sign, but the change in handshape is relatively subtle.

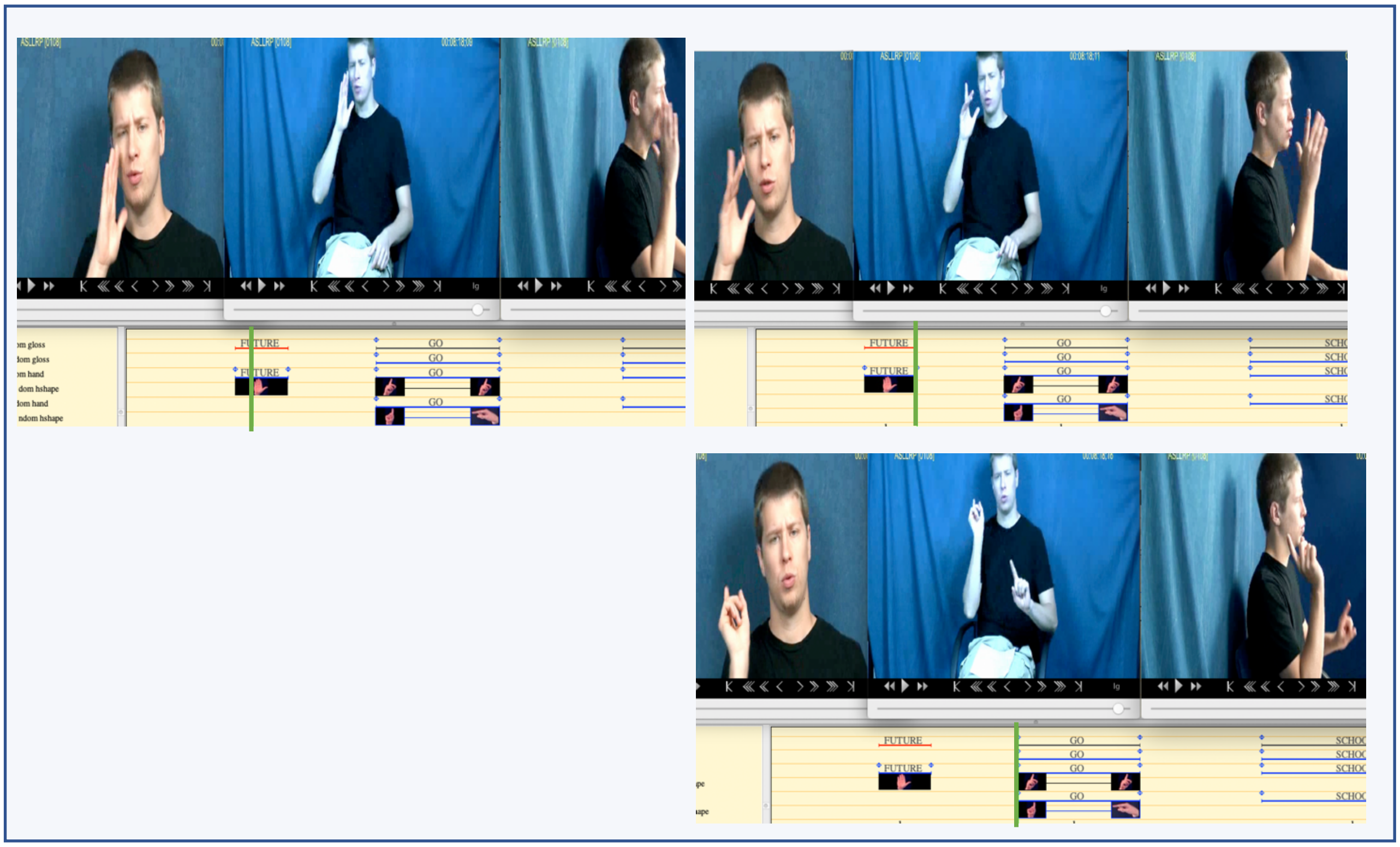

***Figure 7.*** In the upper-right screenshot above, the B-L handshape of FINISH starts to transition, before the end of the sign, to the handshape for the following sign, GO. The green alignment indicator shows the displayed frame.

On the other hand, about 75 (or approximately half) of those 158 cases involve major handshape changes, which would be potential confounds for sign recognition of the segmented sign based on start and end hand configurations. Although these judgments of degree of modification of handshapes are rather subjective, this does give a general idea of the limited extent to which computer-based sign recognition of segmented signs that relies on identification of start and end handshapes of signs would face significant confounds from the effects of handshape coarticulation. In **Figure 8**, the handshapes for HOUSE are shown in the first screenshot, but in the second, showing the final frame of that sign, the dominant handshape has assimilated to the following 1 handshape for the index sign that comes after HOUSE.

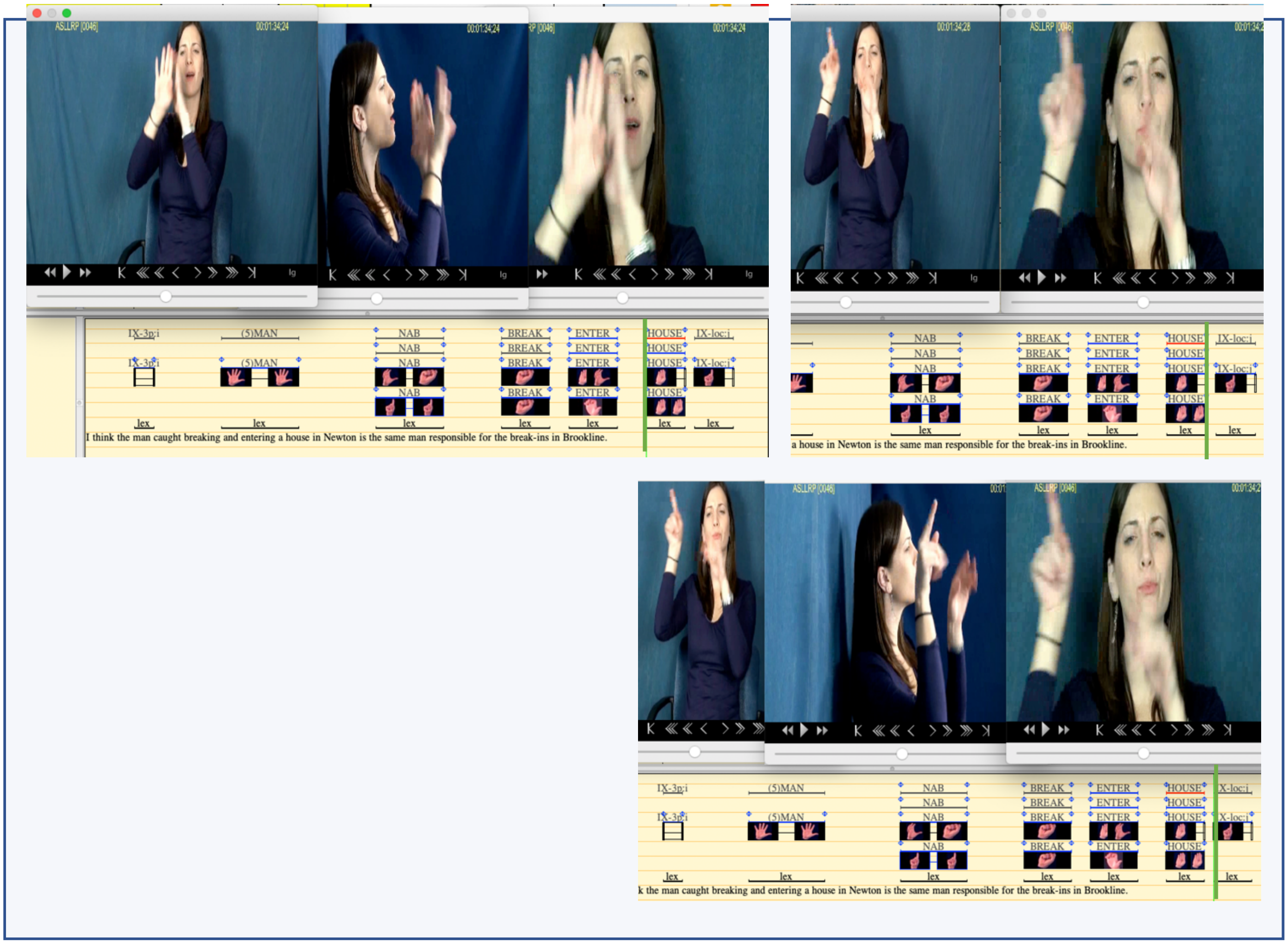

***Figure 8.*** In the upper-right screenshot above, the dominant handshape for HOUSE has assumed the 1 handshape, anticipating the start handshape of the following sign. The green alignment indicator shows the displayed frame.

### 3.1. Prevalence of anticipatory vs. perseverative co-articulation

Cases of regressive coarticulation (where the end handshape of the affected sign is modified to become more like the start handshape of the following sign) occur at more than twice the frequency of progressive coarticulation (where the preceding handshape affects the start handshape of the affected sign).

- Of the 90 cases where 1-handed signs displayed effects of coarticulation, the start handshape was affected by perseverative coarticulation (regressive assimilation) in 29 cases; the end handshape was affected by anticipatory coarticulation (progressive assimilation) in 68 cases; these figures include 7 cases where the sign appeared to display coarticulation with respect to both the preceding and following signs.
- Considering specifically the 47 2-handed signs in which only the dominant hand displayed coarticulation effects, there were 3 cases where the handshapes appeared to display coarticulation with respect to both the preceding and following sign, and a total of 15 instances of perseverative coarticulation, and 35 cases of anticipatory coarticulation.

- Of the 28 2-handed signs in which the non-dominant hand displayed coarticulation effects, there were 5 instances where the start handshapes of both the dominant and the non-dominant hand were affected by the preceding handshapes, and 5 instances where this was the case only for the non-dominant hand. There were 8 instances where the end handshapes of the dominant and non-dominant hand were affected by the following handshapes, and 17 instances where this was the case only for the non-dominant hand.

Another interesting effect involves 2-handed signs in which the dominant hand undergoes coarticulation with a 1-handed sign that precedes or follows: It is sometimes the case for 2-handed signs with the same handshape on both hands that the handshape change on the dominant hand resulting from coarticulation may be exhibited by both hands. For example, in **Figure 9**, the end handshape of the sign on the dominant hand, fingerspelled MARY—the Y handshape—results in a slight modification of the handshape of the dominant hand of the following sign (which would normally be a 5 handshape), but, in fact, this modification is found on both hands of the following sign.

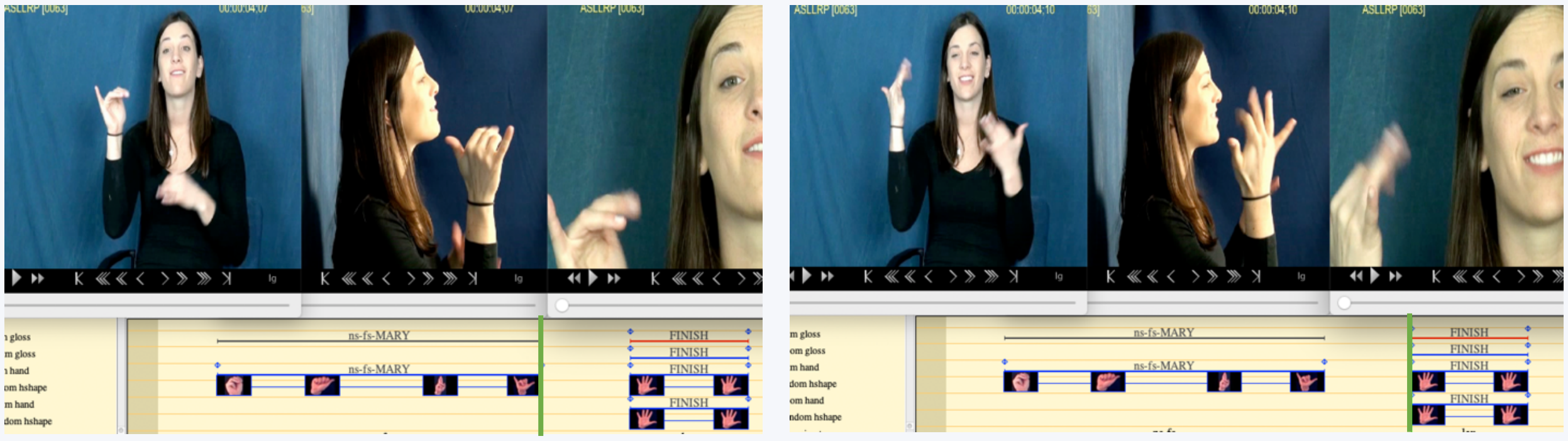

The green alignment indicator marks the displayed frame. Compare the handshapes in the second screenshot above with the unmodified, normal 5 start handshape on both hands for **FINISH**:

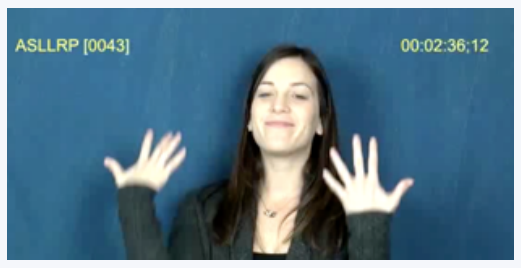

***Figure 9***. Coarticulation triggered by the handshape of the preceding sign on the dominant hand affecting both hands of a 2-handed sign with the same handshape on both hands

### 3.2. Effect of handshape on the likelihood of coarticulation

Discussions of handshape coarticulation in signed languages have generally focused on the 1 and 5 (and similar) handshapes, which are among the most frequently occurring handshapes. One might wonder whether these (or other) handshapes are more likely to undergo coarticulation than other handshapes. If we included all the instances of Index signs, the 1 handshape would certainly be the most susceptible to coarticulation effects. **Excluding those specific lexical items:** coarticulation still occurs relatively frequently with this handshape. The handshape classes that most frequently undergo coarticulation in our dataset are the handshapes shown below (with bent and curved variants included in the relevant classes). Note that those for which there are many instances of coarticulation, relatively speaking, are also those that occur with the greatest frequency overall. Below are listed the total number of cases displaying coarticulation out of the total number of occurrences of the specific handshape classes in the set of data that we analyzed.

- *Class 1:* 1, X, L, D — 24/405 instances: 5.93%

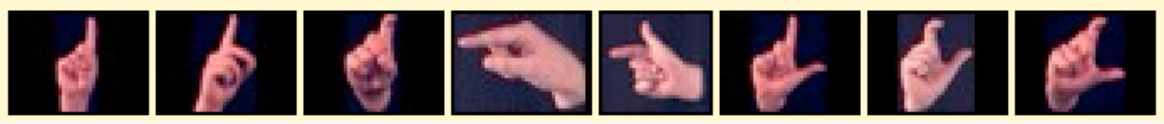

- *Class 2:* 5, B-L, B (including varying positions of the thumb) — 55/992 instances: 5.54%

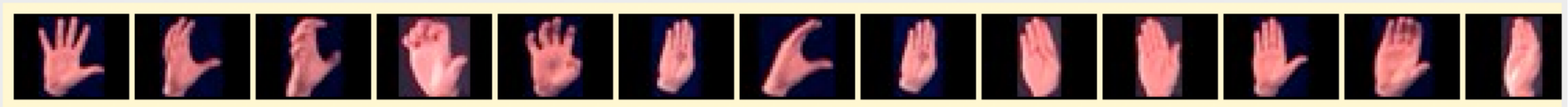

- *Class 3:* O, flat-O — 6/191 instances: 2.91%

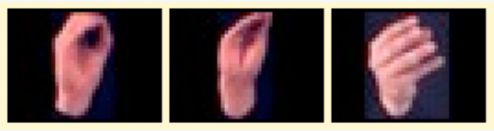

- *Class 4:* A, S, 10 — 11/459 instances: 2.51%

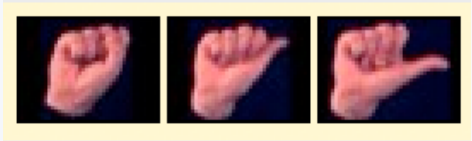

***Table 1.*** Handshape classes with at least 5 examples displaying coarticulation effects: Total number of instances displaying coarticulation effects / total number of instances overall in the data set

There were also 4 examples of coarticulation effects for the Y handshape 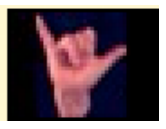 (out of 10 occurrences) and 3 with the P/K handshape 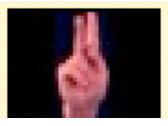 (out of 14 occurrences), but no examples of coarticulation with any of the other handshapes (combined 240 occurrences).

### 3.3. Significance of these findings

Although the judgments about presence or degree of coarticulation effects is somewhat subjective, the overall conclusion—that handshape coarticulation effects are quite limited, with the exception of a few specific lexical items, being a bit more prevalent with some handshapes as compared with others—suggests that computer-based sign recognition from video that attends to start and end handshapes should not be too badly derailed by handshape coarticulation effects.

## 4. Distinguishing 1- and 2-handed Signs: Potential Confounds

Signs may be produced with one or both hands; this is, in general, lexically specified, although there are a few signs that are frequently produced as either 1- or 2-handed, and there are not infrequent exceptional cases where a sign that is normally 1-handed is produced with 2 hands, or vice versa. Much has been written about the role of the non-dominant hand and its phonological representation. Citing Brentari (1990b, 1998), Brentari & Goldsmith (1993), Perlmutter (1991), Sandler (1989, 1993), and van der Hulst (1996), Sandler (2009) states:

> There is a broad consensus that there is only one primary articulator in the lexical phonology of sign language, the dominant hand… This means that the non-dominant hand plays only a minor role in lexical representations. It represents a meaningless phonological element, and its shape and behavior are so strictly constrained as to make it largely redundant.

Nonetheless, since any given sign is typically produced as either 1- or 2-handed, this distinction is potentially useful for computer-based sign recognition from video. However, distinguishing 1 from 2-handed signs is much less straightforward for continuous signing than for citation-form signs.

There are cases in signed languages where there is activity on the non-dominant hand during the production of a 1-handed sign on the dominant hand, which could create the incorrect impression of production of a 2-handed sign. There are also cases where a sign normally produced with one hand is instead produced as 2-handed. Conversely, there are cases where a 2-handed sign is produced as 1-handed: articulated only with the dominant hand. Such phenomena include those listed below.

### 4.1. Perseveration and Anticipation

Sometimes the handshape on one hand (most frequently, the non-dominant hand) of a 2-handed sign remains in place during the articulation of one or more 1-handed signs on the other hand (or it gradually transitions to the handshape required on that hand for the next 2-handed sign; see Section 4.2). This has been referred to as "H2 spread" in Nespor & Sandler (1999), and it does not extend beyond the boundaries of the phonological phrase. It has also been referred to as "weak hand hold" and analyzed in various ways. An example from the Sign Language of the Netherlands, NGT, presented in Crasborn (2011) "The Other Hand in Sign Language Phonology," p. 233, is shown in **Figure 10**.

**Figure 11** shows another example from the ASLLRP SignStream® 3 corpus (Cory_2013-6-27_sc107.ss3, Utterance 35). In this latter case, the non-dominant handshape from the sign DRIVE is held while the following two 1-handed signs are produced.

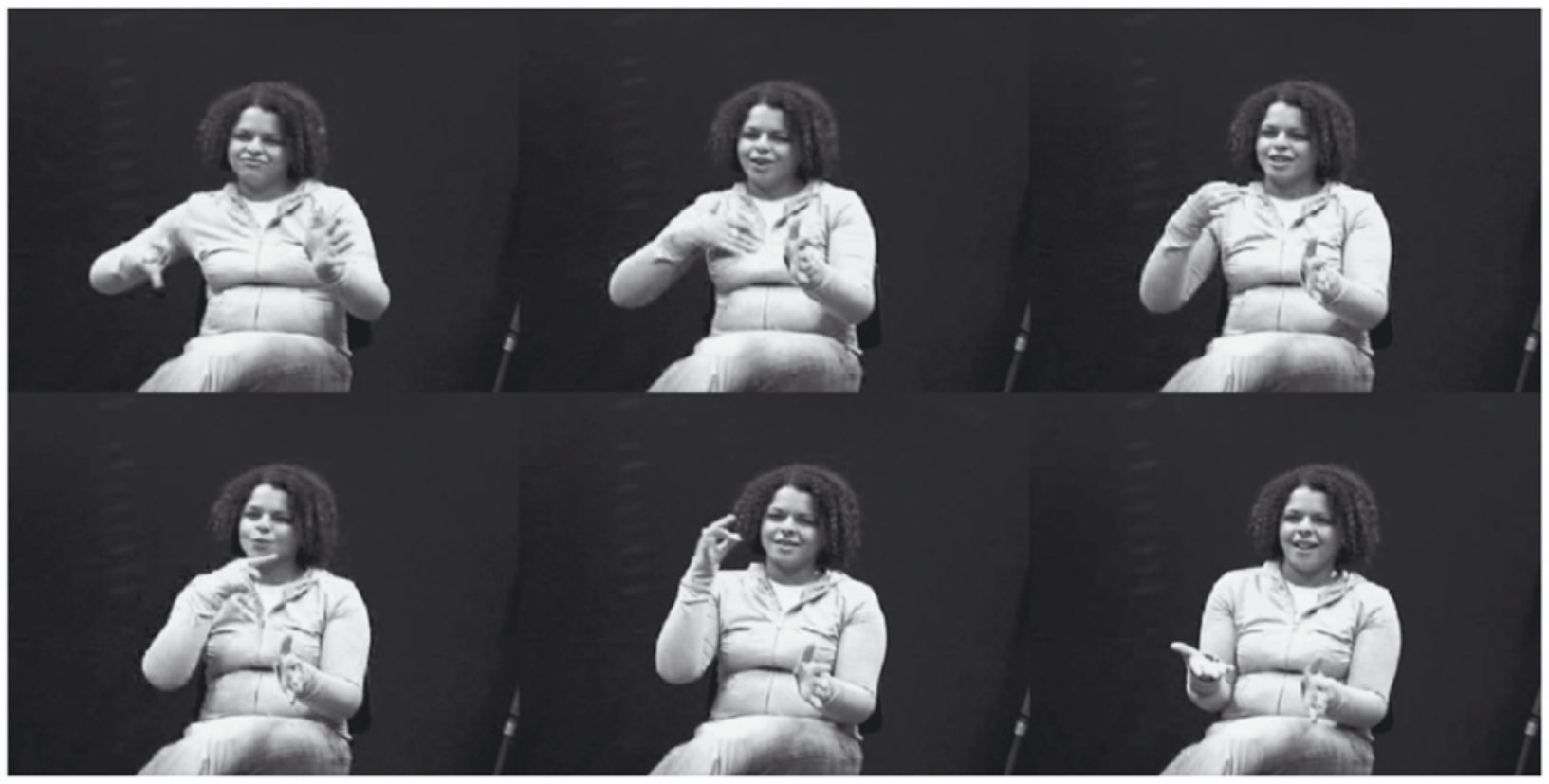

Figure 10.7 Non-dominant hand spreading marks the size of a prosodic domain. Source: Corpus NGT, Session 62 (01:13–01:16). The non-dominant hand holds the two-handed item TO-SIGN, while the dominant hand continues to sign FEEL AT-HOME MOTHER^TONGUE PALM-UP. 'Signing feels like you're using your mother tongue'

**Figure 10**. Example of non-dominant hand spread in NGT, from Crasborn (2011)

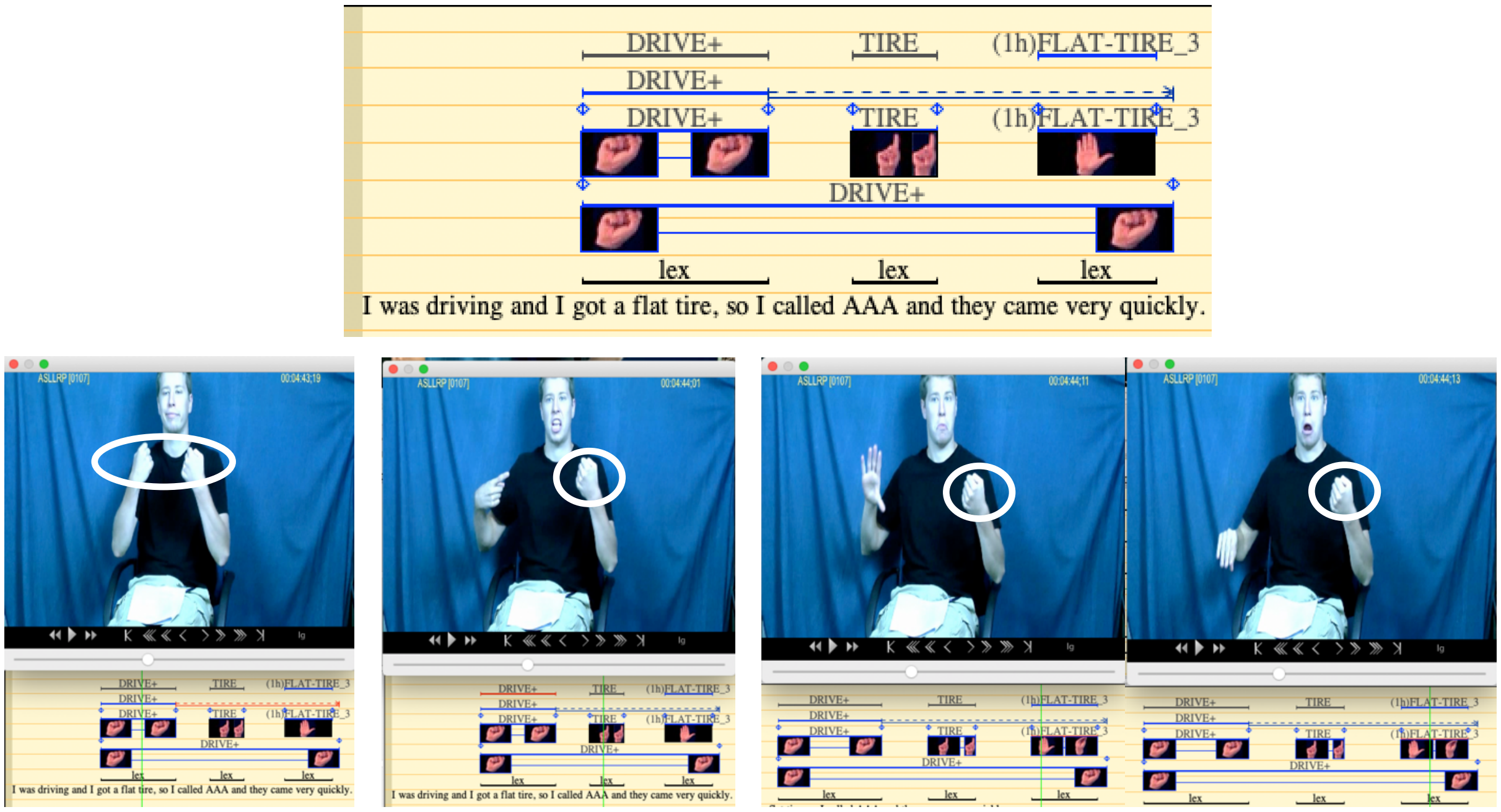

Dominant hand:

DRIVE TIRE (1h)FLAT-TIRE (start; end)

Non-dominant hand:

DRIVE________________________________________________________________

**Figure 11**. Example of non-dominant hand spread in ASL

When there are two 2-handed signs with similar non-dominant handshapes used for both, separated by one or more 1-handed signs, it is quite common for the non-dominant handshape to remain in place during the 1-handed sign(s), as illustrated in **Figure 12** from the ASLLRP SignStream® 3 corpus (Rachel_2011-12-08_sc44.ss3, Utterance 3).

Sometimes during a 1-handed sign, the hand that is not used for that sign gets into position for the next 2-handed sign a bit in advance.

In all of these cases, it can appear that the non-dominant hand is participating in the production of what is, in fact, a 1-handed sign, i.e., where the production could be mistaken for a 2-handed sign.

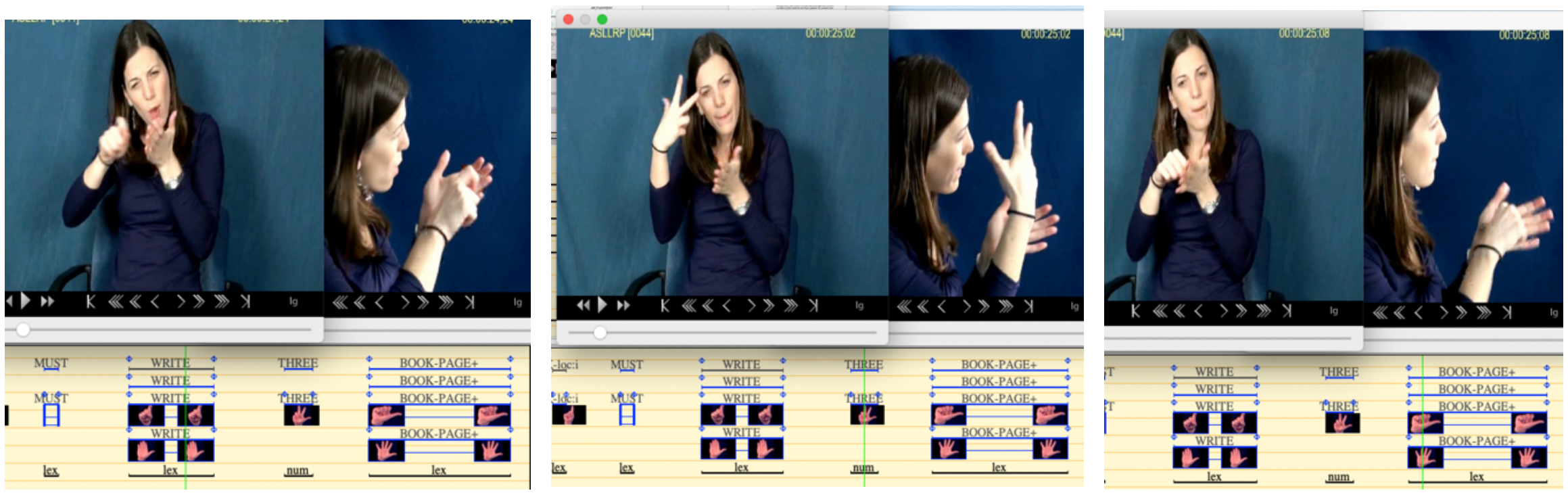

(A) WRITE (B) **THREE** (1-handed sign) (C) BOOK-PAGE

**Figure 12.** Non-dominant hand remaining in place during production of a 1-handed sign

### 4.2. Mirroring

Sometimes a 1-handed sign is produced with the other hand mirroring the action of the dominant hand. See **Figure 13** for some examples of mirroring from Nilsson (2007), "The Non-Dominant Hand in a Swedish sign Language Discourse," pp. 10-11.

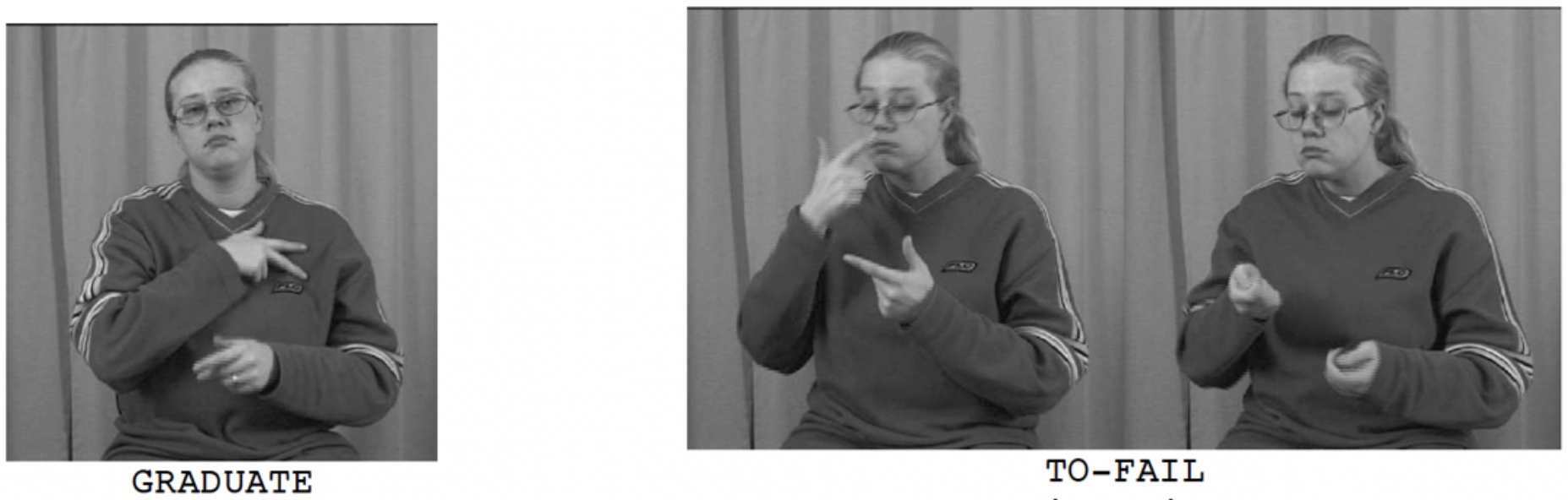

GRADUATE TO-FAIL

**Figure 13**. Examples of mirroring from Swedish Sign Language (from Nilsson, 2007); these signs are normally 1-handed

This happens frequently with index signs, e.g., and it should be noted that in this case, the index fingers on the two hands are likely to point in different (mirror-image) directions. An example of that from the ASLLRP SignStream® 3 corpus is shown in **Figure 14**, with mirroring of the index sign that is highlighted.

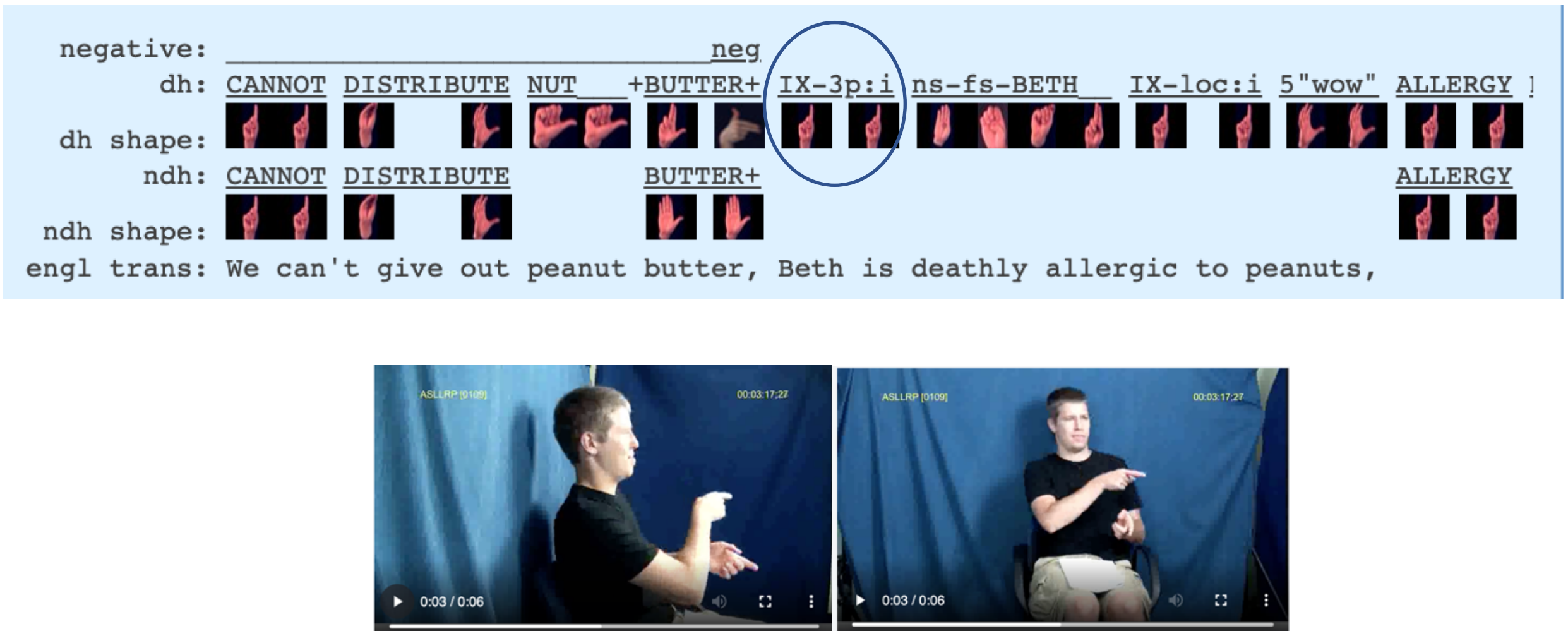

**Figure 14.** Examples of mirroring of an index sign, from the ASLLRP SignStream® 3 corpus (Cory_2013-6-27_sc109, Utterance 26)

### 4.3. Independence of signing on the 2 hands

Sometimes you may have 1-handed signs produced independently by each of the 2 hands overlapping partially or totally. One example from the ASLLRP SignStream® 3 corpus (2-Ben-Voice-Identity.ss3, Utterance 14) is shown in **Figure 15**. The sign PHONE, produced on the non-dominant hand is held, such that it overlaps with the sign for EXPERIENCE subsequently produced by the dominant hand.

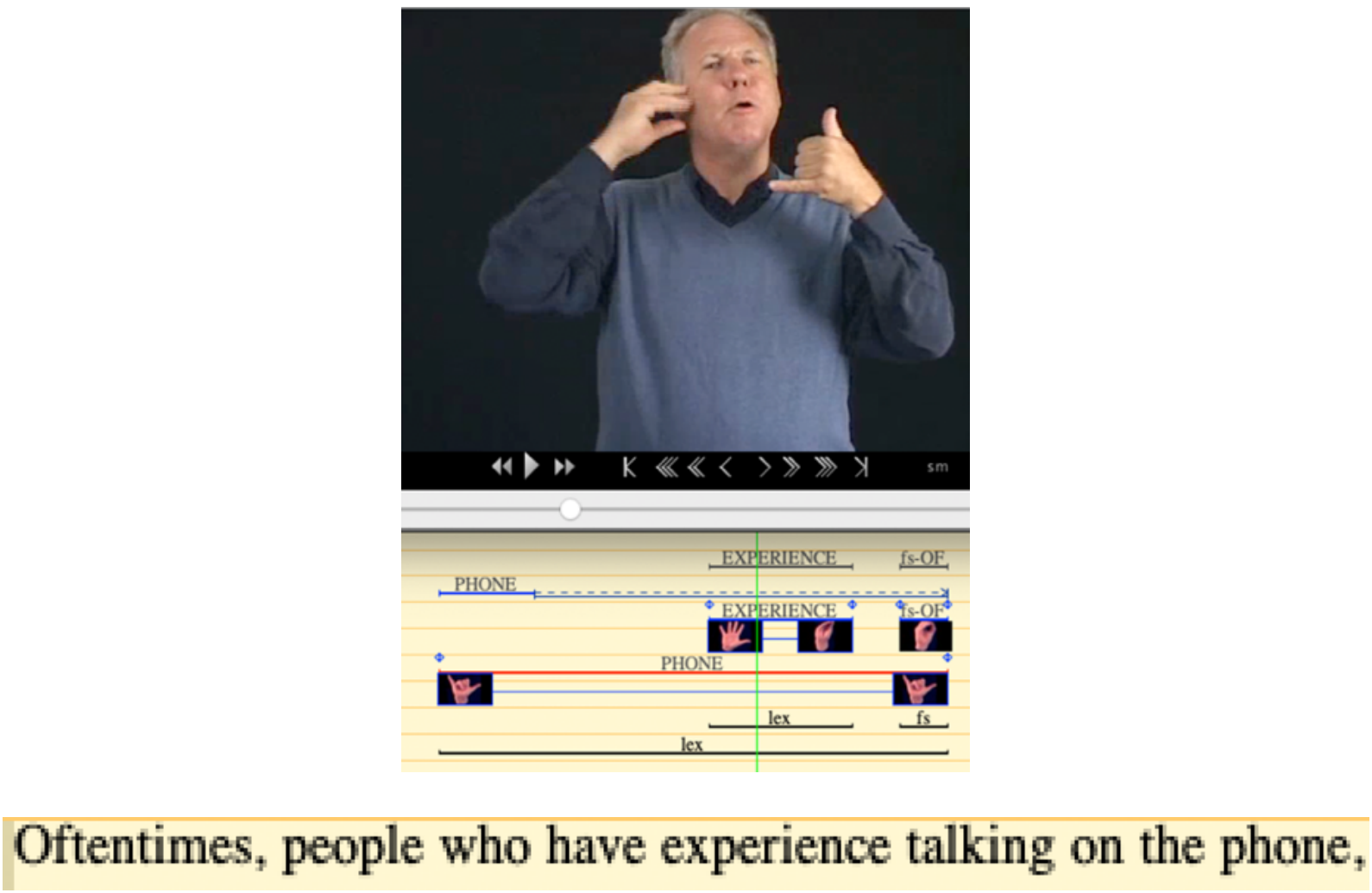

**Figure 15.** Example of 2 independent 1-handed signs produced on the 2 hands

It can also happen that either of the two hands producing separate signs may be producing 1-handed versions of signs that are normally 2-handed. In **Figure 16**, from the ASLLRP SignStream® 3 corpus (2-Ben-Voice-Identity.ss3, Utterance 2), the 1-handed sign glossed as TALKwg, produced with the dominant hand, overlaps with a 1-handed version of the normally 2-handed sign for PEOPLE (see Section 4.6).

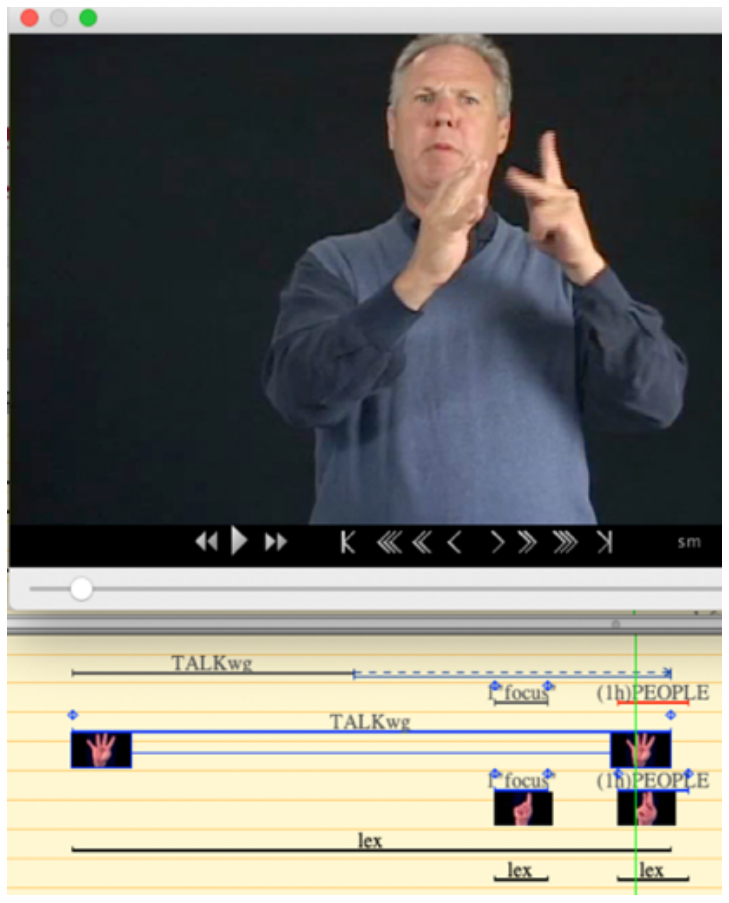

When a person hears the speech of another, they can tell who that person is and the identity of that person.

**Figure 16.** A 1-handed sign on the dominant hand overlapping with a 1-handed version of a normally 2-handed sign on the non-dominant hand

One situation in which there are important things happening separately on the two hands, but in a way that is interrelated, involves classifier constructions. It is frequently the case that two different classifiers can be produced on the two hands, to establish a relationship between them. One example is shown in **Figure 17**, taken from the National Center for Sign Language & Gesture Resources (NCSLGR) SignStream® 2 Corpus (Neidle*, et al.* 2022b) (three pigs.xml, Utterance 24), where the non-dominant hand represents a base surface, and the dominant hand acts out spreading mortar over that surface.

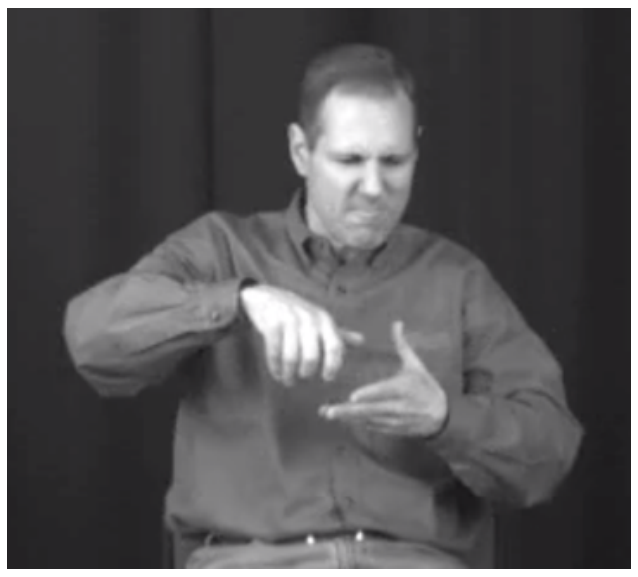

```
            main gloss: MAKE  ICL:5"spreading mortar" MAKE  ICL:5"spreading mortar":1 BUIL
non-dominant hand gloss:        DCL:5"base surface"_____       DCL:open-B"base surface"
```

**Figure 17**. Different classifiers on the two hands

### 4.4. Focus marker

In some cases, the non-dominant hand points to what is being signed by the dominant hand, to draw attention to it. This happens especially with fingerspelling. We have glossed this sign as '1"focus".' **Figure 18** shows an example from the ASLLRP SignStream® 3 corpus (1-Ben-Introduction.ss3, Utterance 15, and Rachel_2012-02-14.ss3, Utterance 66).

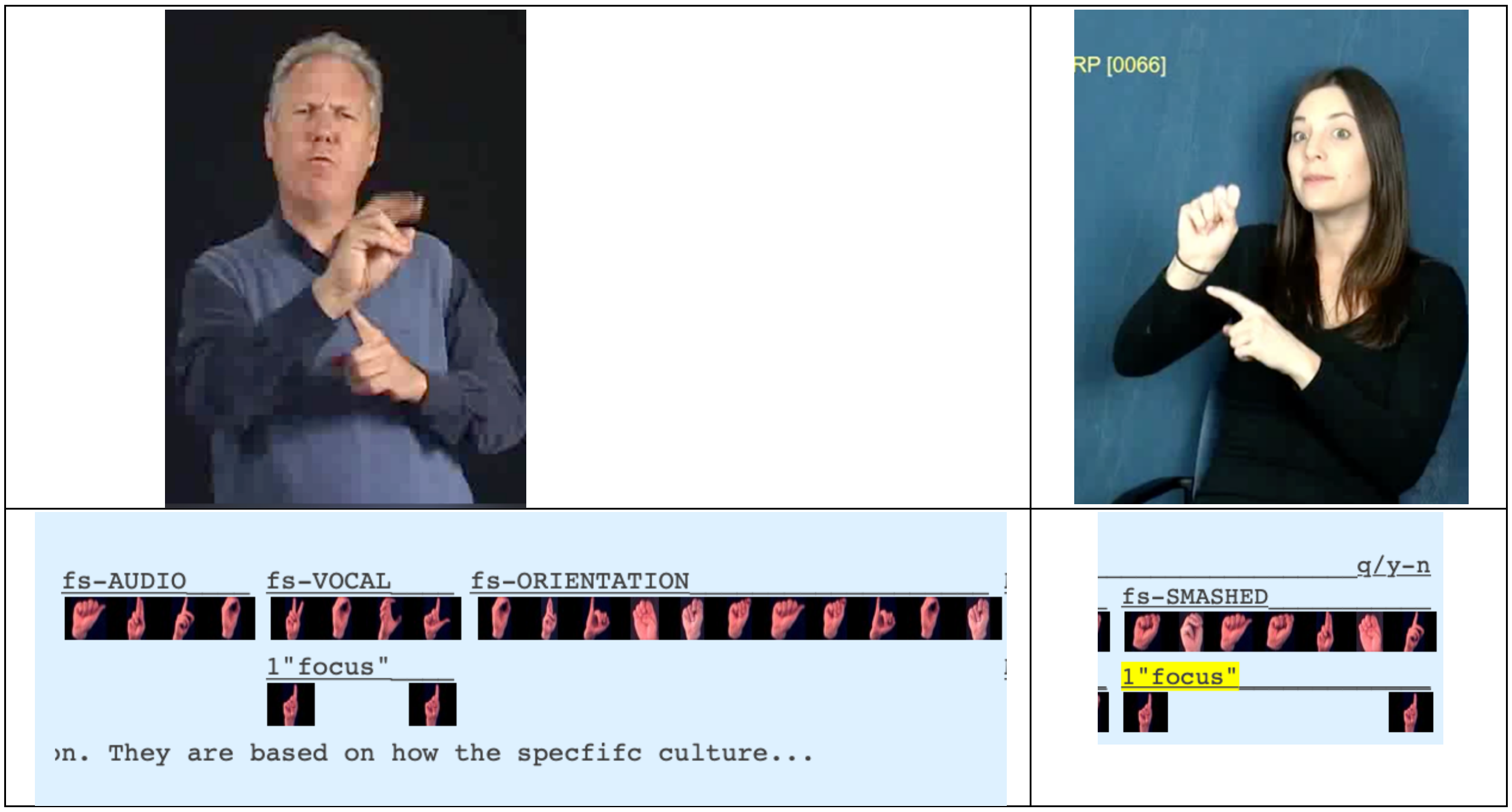

**Figure 18**. Focus marker on the non-dominant hand used especially with fingerspelling

### 4.5. Buoys

Liddell (2003), p. 223, defines buoys as follows:

> Signers frequently produce signs with the weak hand that are held in a stationary configuration as the strong hand continues producing signs. Semantically they help guide the discourse by serving as conceptual landmarks as the discourse continues. Since they maintain a physical presence that helps guide the discourse as it proceeds I am calling them buoys. Some buoys appear only briefly whereas others may be maintained during a significant stretch of signing.

In particular, he defines the 'theme' buoy (p. 242) as "a raised, typically vertical index finger on the weak hand held in place as the strong hand produces one or more signs." An example from the ASLLRP SignStream® 3 corpus (5-Ben-Grunts.ss3, Utterance 2) is shown in **Figure 19**.

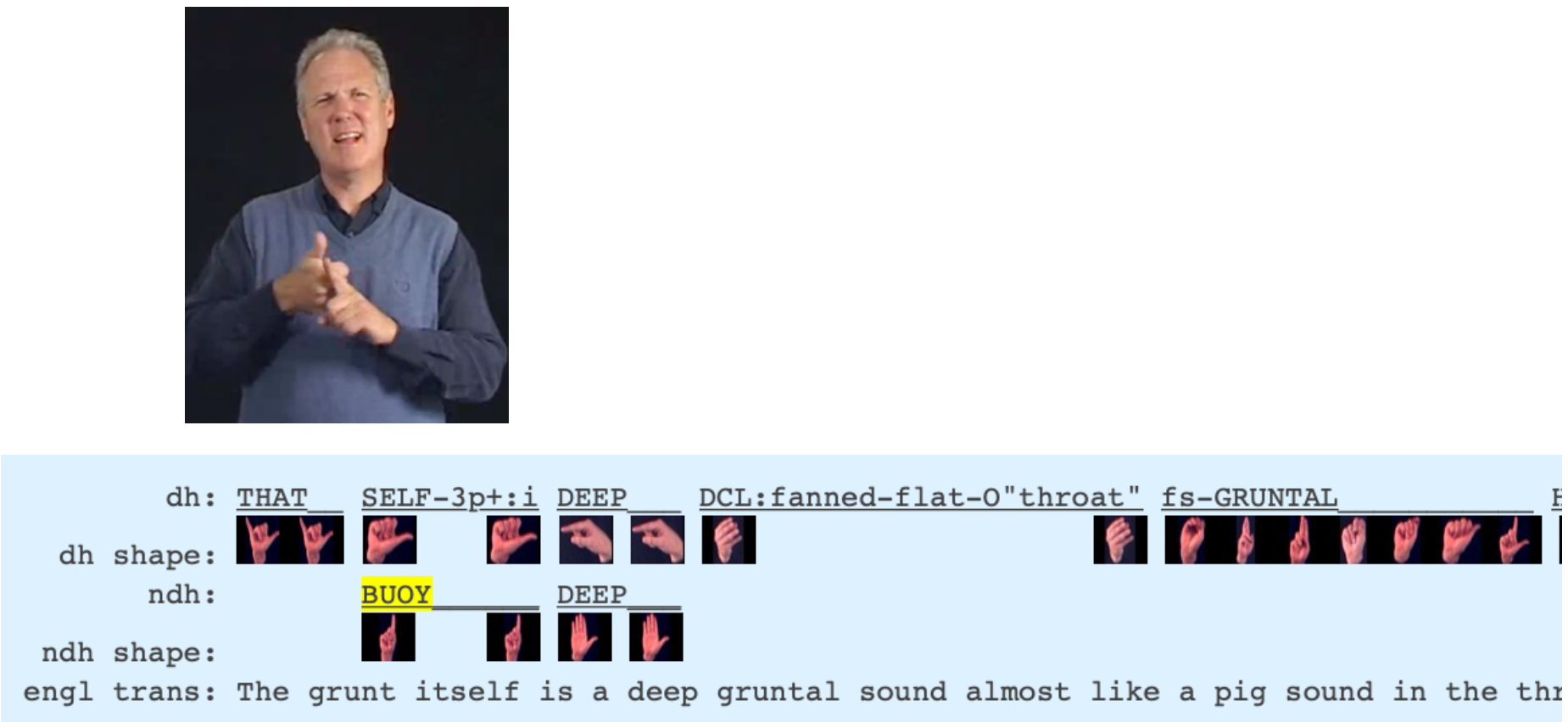

**Figure 19**. Theme 'buoy' on the non-dominant hand occurring with SELF on the dominant hand

### 4.6. Marked number of hands

Sometimes a sign (with the appropriate properties) that is normally produced as 2-handed may be realized with only one hand, a phenomenon sometimes referred to as "weak drop," following Padden & Perlmutter (1987). Compare the examples from the ASLLRP SignStream® 3 corpus shown in **Figure 20**. It also happens that a sign normally produced as 1-handed may be produced with the other hand doubling the properties of the dominant hand, as shown in **Figure 21**. Following our glossing conventions, we attach a prefix—either (1h) or (2h)—to the gloss label for a sign produced with a marked number of hands.

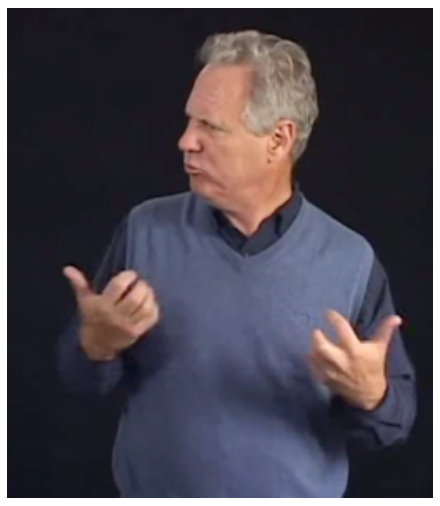

ANGRY

7-Ben-Social-Msg.ss3, Utterance 29

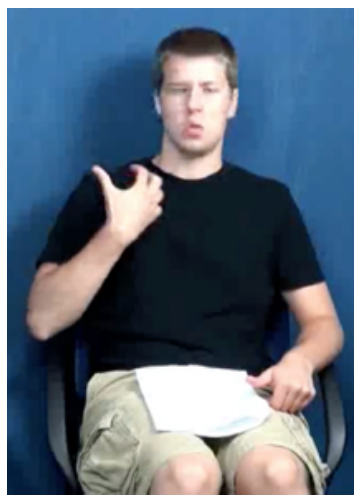

(1h)ANGRY

*Cory_2013-6-27_sc111.ss3, Utterance 21*

**Figure 20**. The normally 2-handed sign ANGRY can exceptionally be produced with 1 hand

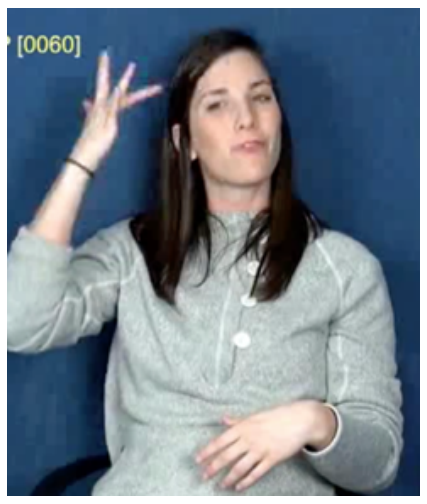

BALD

Rachel_2012-02-09_sc60.ss3, Utterance 13

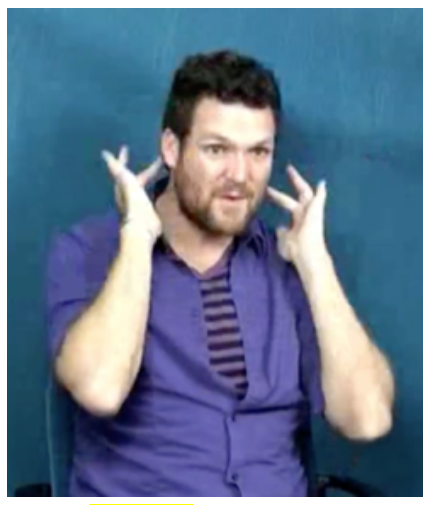

(2h)BALD

Jonathan_2012-11-27_sc93.ss3, Utterance 8

**Figure 21.** The normally 1-handed sign BALD can exceptionally be produced with 2 hands

In sum, there are quite a few cases where computer-based sign recognition will face a challenge in determining whether a given production on one hand in a continuous sign sequence is essentially a 1-handed sign or part of a 2-handed sign, based solely on perceived activity on the two hands. Fortunately, there are other cues that can be exploited to identify whether what is signed on the dominant hand is a 1-handed sign, or whether it is part of a 2-handed sign.

## 5. Distinguishing 1- and 2-handed signs: Exploiting regularities in the linguistic structure of 2-handed signs

Battison (1974) formulated two conditions on ASL sign well-formedness, reformulated in Battison (1977, 1978), which have subsequently been emended slightly by other researchers (e.g., Channon 2004; Eccarius & Brentari 2007); see also discussion in Crasborn (2011)). These have subsequently been observed to hold in other sign languages and are generally now presumed to be universal.

(1) The Symmetry Condition (from Battison (1977, 1978))

> If both hands of a sign move independently during its articulation, then both hands must be specified for the same location, handshape, and movement (whether simultaneous or in alternation), and the specifications for orientation must be either symmetrical or identical.

(2) The Dominance Condition (as paraphrased in van der Hulst & van der Kooij (2021))

> In 2-handed signs in which one hand (the passive or weak hand) is the location for the other (active or strong) hand, the location hand either has the same handshape as the active hand or is selected from a small set of 'unmarked' handshapes.

Note that the determination of the "same" handshape on both hands obviously does not imply that the configuration on the 2 hands is identical, as there is inevitably some phonetic variability in the actual configurations that hands can assume for each linguistically established handshape.

What this means, in particular, is that at least some of the cases discussed in Section 4 where a 1-handed sign is articulated by the dominant hand, but where it appears that both the dominant and non-dominant hands are both active, especially instances where two independent signs are produced on the two hands, could be excluded as potential 2-handed signs because they would not constitute well-formed 2-handed signs, based on combined information about handshapes and movement patterns on the two hands.

We have a large database with annotations of start and end handshapes for the dominant and non-dominant hands (Neidle*, et al.* 2022b), obviously also including information for each sign production about whether it is 1- or 2-handed, which can be used for training; so the computational model should be able to learn to identify 1- and 2-handed signs. The presence of these regularities means that this distinction should be learnable.

## 6. Other Ways in which Linguistic Properties of Signs can be Exploited to Boost Sign Recognition

### 6.1. Segmentation of continuous sign sequences based on sign types

Signed languages differ from spoken languages in that "words" have different internal composition based on the morphological classes to which they belong. Whereas in spoken languages, meaningful units are made up out of linear sequences of phonemes (subject to a variety of phonological processes), different types of signs have different internal composition. This means that one cannot apply a uniform sign recognition strategy.

**Fingerspelled** signs, used principally for borrowings from spoken language and proper nouns, are spelled out using the manual alphabet. Furthermore, fingerspelled signs frequently have letters missing and/or transposed; the first letter is most reliable, and the last letter is the next most reliable. Thus, there is no fixed vocabulary of possible fingerspelled signs. **Lexical** signs, composed of combinations of particular handshapes, orientations, locations (places of articulation), and movements, and in some cases distinctive nonmanual expressions, are the most productive class. **Loan signs** are **lexicalized fingerspelled signs**: signs that originated as fingerspelled, but taken on characteristics of lexical signs, such as specific location, orientation, and/or movement, while often losing some of their component letters. **Classifier constructions** are especially challenging for computer-based recognition. They are morphologically complex and, although they often contain components that are conventional, they incorporate various types of iconicity in representing size, shape, and movement. See, e.g., Benedicto & Brentari (2004); Brentari (2019); and Emmorey, ed. (2003) *Perspectives on Classifier Constructions in Sign Languages.* **Gestures** are also only somewhat conventional in their production.

In order to tailor sign recognition to the appropriate class of signs, one approach is to first segment the continuous signing into segments containing specific classes of signs. This is possible, in principle, because these different types of signs have different visual properties, although there are a small number of cases where the sign production is potentially consistent with more than one class. For example, the loan sign #AC has handshape and movement properties consistent with fingerspelled or lexical or loan signs. However, Yanovich, *et al.* (2016) "Detection of Major ASL Sign Types in Continuous Signing for ASL Recognition" succeeded in accurately matching manual annotations of sign type for over 91% of the video frames from 500 continuous utterances from the publicly accessible NCSLGR corpus (Neidle*, et al.* 2022b), including 7 signers. They used a multiple learning-based segmentation method that exploited both motion and shape statistics of regions of high local motion.

### 6.2. Boosting handshape recognition

Accurate recognition of handshapes is important for sign recognition, but recognizing 3D handshapes from 2D video is challenging. Linguistic constraints on the relationships between the handshapes on the two hands for 2-handed signs, and between the allowed start and end handshape combinations in lexical signs specifically, can serve to boost the accuracy of handshape recognition, and by extension, of sign recognition.

➢ ***Handshape recognition in 2-handed signs***

The handshape, orientation, and movement properties of the sign can be assessed in relation to the Symmetry and Dominance Conditions, to determine the sign type. A chart showing sign types based on Battison (1978) is shown in **Figure 22**.

Once the sign has been categorized, e.g., as having essentially the same (see the caveat on p. 20) or different handshapes on the two hands, taking account of the constraints that apply to that sign type can enhance sign recognition. For 2-handed signs that have essentially the same handshape on both hands from the set of handshapes that are used for linguistic purposes in ASL, the visual information from both hands can be combined to arrive at the maximally probable handshapes. For sign classes in which the handshapes are different on the two hands, with the non-dominant handshape restricted to a small set of possible handshapes, that limitation can also boost recognition accuracy of the non-dominant handshape.

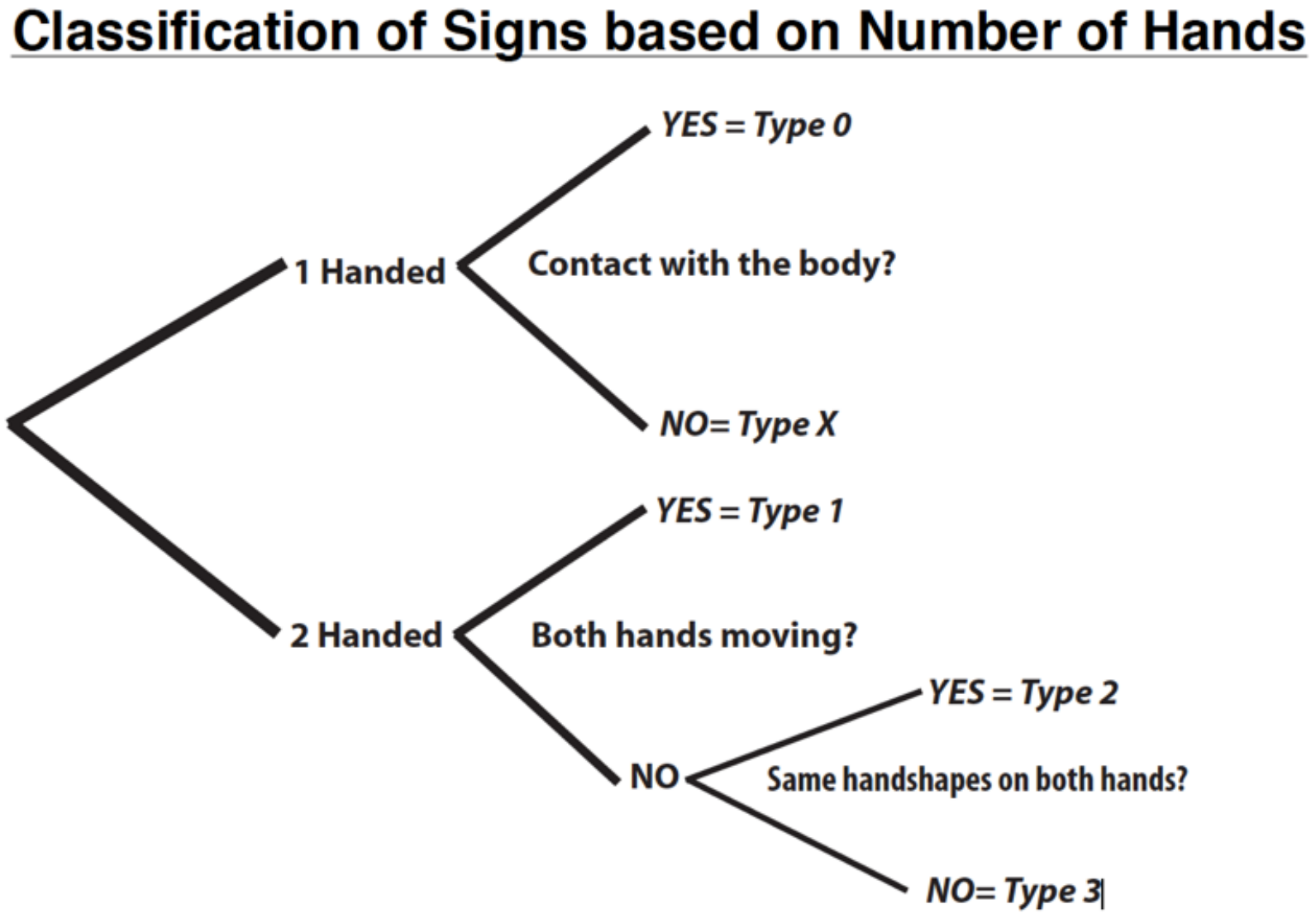

| Sign Type | Description | Notes |
|---|---|---|
| **Type 0** | One handed, **not** contacting the body | |
| **Type X** | One handed, contacting the body (but **not** other hand) | **Types 0 and X** account for ~ 40% of signs |
| **Type 1** | Two handed, both moving, synchronous or alternating movements | **Type 1** accounts for ~ 35% of signs |
| **Type 2** | Two handed, one active, one passive, **same** handshape | **Types 2 and 3** account for ~ 25% of signs |
| **Type 3** | Two handed, one active, one passive, **different** handshape | The handshape of the **non-dominant** hand is unmarked (5,1,B,A,C or O) |

**Figure 22.** Sign types based on Battison (1978)

- ***Constraints on start/end handshape pairs in lexical signs***

It has been observed by many researchers that, for monomorphemic lexical signs, the change from the start hand configuration to the end hand configuration is severely constrained, although the specifics of the theoretical account have differed a bit (e.g., Sandler 1989; Brentari 1990a; Liddell 1990). Handshape recognition can be boosted by taking into account the likelihood of specific start/end handshape combinations. In previous research, these probabilities, based on the statistics from our corpora, have been exploited to boost handshape recognition (Thangali*, et al.* 2011; Neidle*, et al.* 2012; Thangali 2013; Dilsizian*, et al.* 2014).

For example, in our ASLLVD dataset (see Neidle (2012; 2022a; 2022b)), the three most common dominant hand start handshapes are shown in column 1 of **Figure 23**, in decreasing order of frequency.[5] Column 2 shows the most likely end handshape given the start handshape in the same row, and other attested end handshapes are shown in the remaining columns of that row, again in decreasing order of frequency. The numbers below each handshape label represent the total number of occurrences in our dataset. A magnified version of the first part of this chart is displayed in **Figure 24**.

So, of the 1,089 examples that start with the B-L handshape, 917 of them also end with that same handshape, 89 end in the bent-B-L handshape, and so on. Not surprisingly, for most of the start handshapes, the most likely end handshape is that same handshape, i.e., with no change in handshape from the start to end of the sign. It is clear that the frequencies drop quickly, with only a few of the 87 linguistically distinguished handshapes being at all likely as end handshapes, and the entire set of attested end handshapes for a given start handshape is a small fraction of the 87 possible handshapes. In this case, only 4 other end handshapes occur with >1% likelihood, based on these statistics.

[5] The full charts of overall and co-occurrence handshape frequencies for our ASLLVD and ASLLRP continuous signing corpora are available from <https://dai.cs.rutgers.edu/dai/s/runningstats>

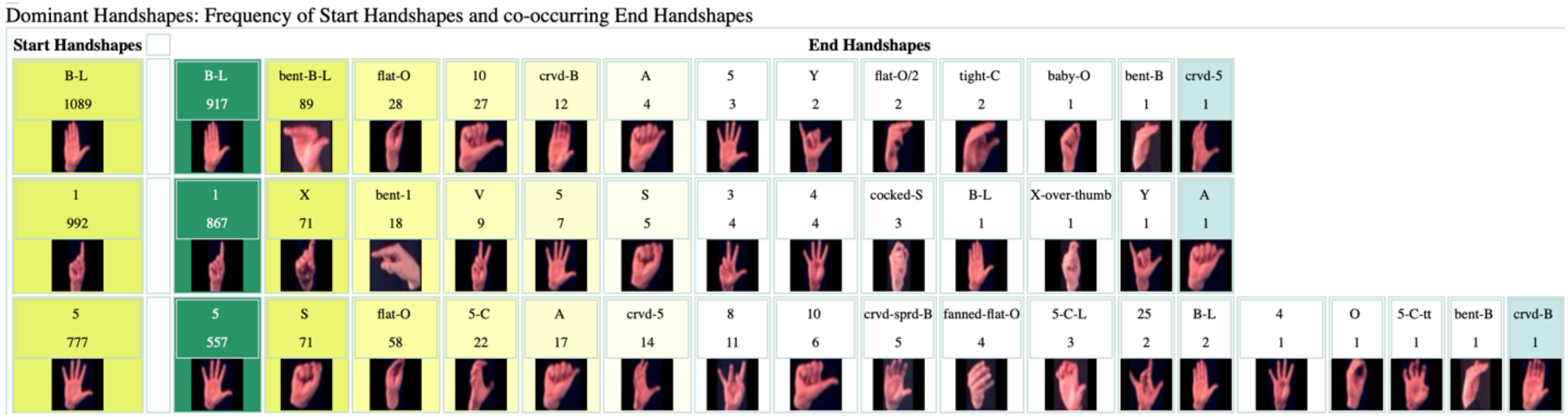

***Figure 23.*** Frequency of start handshapes and start/end handshape pairs from the ASLLVD

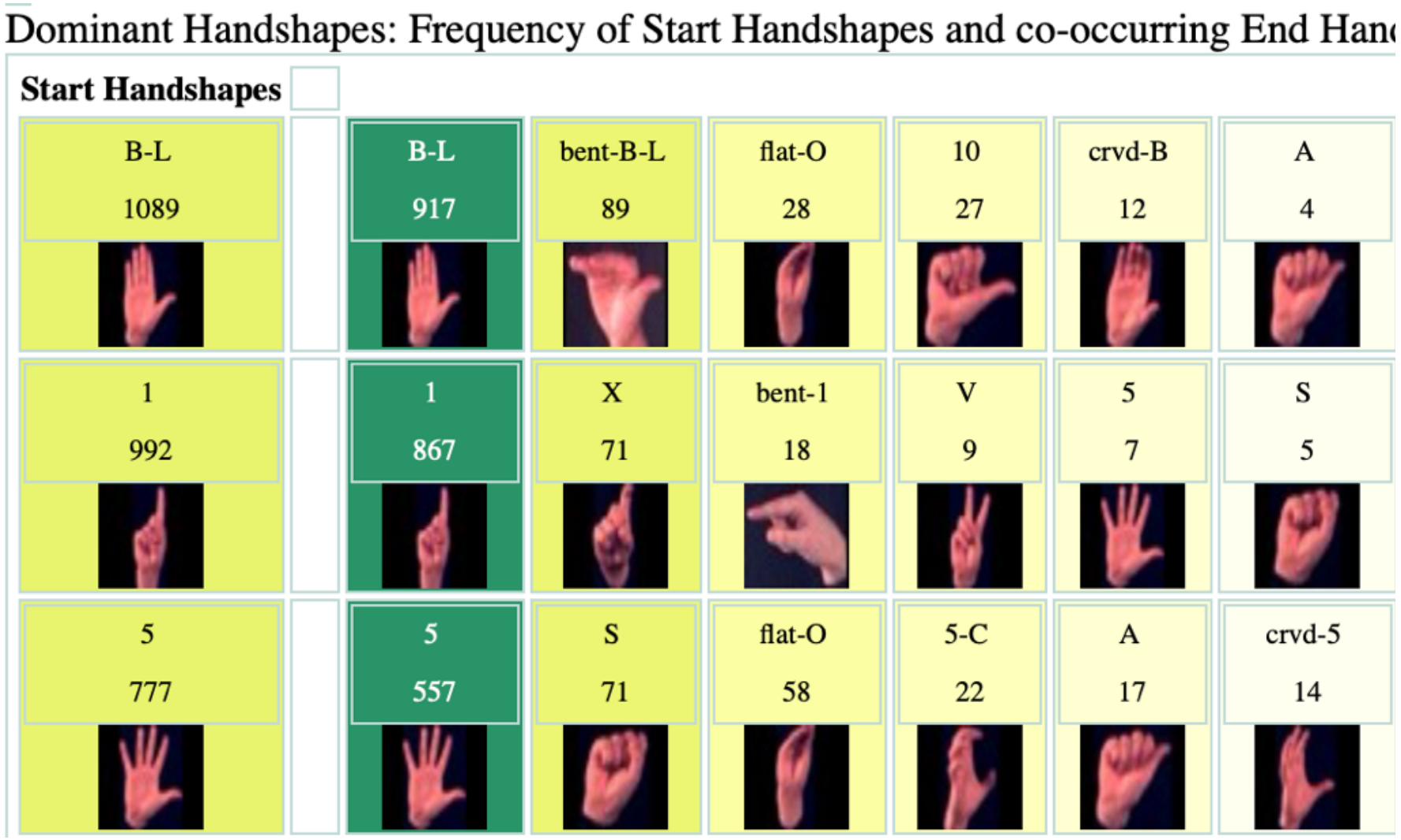

**Figure 24.** Magnified view of the first part of **Figure 23**

In Thangali, *et al.* (2011), which utilized a HandShapes Bayesian Network (HSBN) that modeled the transition probabilities between start and end handshapes in monomorphemic lexical signs, using the statistics on start/end handshape combinations increased rank-1 handshape recognition accuracy by about 46%—from 30.4% (597 of 1962) to 44.4% (871 of 1692)—using a subset of the ASLLVD that included 1500 ASL lexical signs, where each lexical sign was produced by three native ASL signers. In Dilsizian, *et al.* (2014), which used a method to estimate 3D hand configurations to distinguish among 87 hand configurations linguistically relevant for ASL, handshape recognition accuracy was improved by by 15.1%, from 71.02% to 81.76%, through incorporation of statistical information about the linguistic dependencies among handshapes within a sign, derived from the annotated ASLLVD corpus of almost 10,000 sign tokens.

## 7. Conclusions

We have attempted here to discuss the various factors that present a challenge for computer-based recognition of signs from natural, continuous signing that make this a more difficult task than recognition of isolated, citation-form signs. We had anticipated that coarticulation would be much more prevalent than it has proved to be in the data we analyzed, although there are significant coarticulation effects. We also considered here other factors that present special challenges for distinguishing and recognizing 1-handed signs (including cases where there are two overlapping signs being produced separately on the two hands) and 2-handed signs in continuous signing. Furthermore, we discussed several linguistic constraints as they apply specifically to lexical signs, which can be leveraged to boost handshape recognition and to improve sign recognition more generally.

## *Acknowledgments

This research would not have been possible without the excellent work on: development of SignStream® software for linguistic annotation of sign language data (thanks especially to Gregory Dimitriadis, the principal SignStream® developer); development of our Web-based database system for providing access to the linguistically annotated video data (thanks especially to Augustine Opoku); invaluable help from many people for linguistic annotation of the data (thanks especially to Carey Ballard and Indya Oliver); and collaboration with Dimitris Metaxas and his students and colleagues at Rutgers University (especially Konstantinos Dafnis) on computer-based sign language recognition from video. This work also benefited from discussions with BU graduate student Neil Ray.

We are grateful to the ASL signers who provided video data (those shown here include: Ben Bahan, Corey Behm, Rachel Benedict, Elizabeth Cassidy, Lana Cook, Jonathan McMillan, and Braden Painter). We are also very grateful to Matt Huenerfauth at RIT for overseeing some data collection of continuous signing at RIT, and also to Joe Dannis and Becky Ryan from DawnSignPress for sharing wonderful video data with us.

This work was supported in part by grants # 2235405, #2212302, 2212301, 2212303 from the National Science Foundation, although any opinions, findings, and conclusions or recommendations expressed in this material are those of the author and do not necessarily reflect the views of the National Science Foundation.

## Table of Contents